\documentclass[10pt,twocolumn,letterpaper]{article}

\usepackage[pagenumbers]{cvpr} 

\usepackage{graphicx}
\usepackage{amsmath}
\usepackage{amssymb}
\usepackage{booktabs}
\usepackage{multirow}
\usepackage{float}
\usepackage{tabularx} 
\usepackage{ragged2e} 
\usepackage{booktabs} 
\usepackage[accsupp]{axessibility}
\usepackage[pagebackref,breaklinks,colorlinks]{hyperref}

\usepackage[capitalize]{cleveref}
\crefname{section}{Sec.}{Secs.}
\Crefname{section}{Section}{Sections}
\Crefname{table}{Table}{Tables}
\crefname{table}{Tab.}{Tabs.}

\def\cvprPaperID{1791} 
\def\confName{CVPR}
\def\confYear{2023}

\begin{document}

\title{Exploring Deep Models for Practical Gait Recognition\thanks{
We express our sincere gratitude for the valuable feedback provided by reviewers, despite the multiple rejections this paper has faced. Recognizing its imperfections, we have made the decision not to resubmit this manuscript. However, in response to numerous code requests, we are committed to sharing the code associated with this work. Should this code prove beneficial to your research endeavors, we kindly request your acknowledgment through citation. Your support is greatly appreciated.
}}

\author{
Chao Fan\textsuperscript{\rm 1}, Saihui Hou\textsuperscript{\rm 2,\rm 3}, Yongzhen Huang\textsuperscript{\rm 2,\rm 3}, and Shiqi Yu\textsuperscript{\rm 1}\thanks{Corresponding Author} \\ 
{\normalsize $^1$ Southern University of Science and Technology}\\
{\normalsize $^2$ Beijing Normal University} {\normalsize $^3$ WATRIX.AI} \\
{\tt\scriptsize 12131100@mail.sustech.edu.cn, \{housaihui, huangyongzhen\}@bnu.edu.cn, yusq@sustech.edu.cn}
}

\maketitle
\begin{abstract}
Gait recognition is a rapidly advancing vision technique for person identification from a distance. 
Prior studies predominantly employed relatively shallow networks to extract subtle gait features, achieving impressive successes in constrained settings. 
Nevertheless, experiments revealed that existing methods mostly produce unsatisfactory results when applied to newly released real-world gait datasets. 
This paper presents a unified perspective to explore how to construct deep models for state-of-the-art outdoor gait recognition, including the classical CNN-based and emerging Transformer-based architectures. 
Specifically, we challenge the stereotype of shallow gait models and demonstrate the superiority of explicit temporal modeling and deep transformer structure for discriminative gait representation learning. 
Consequently, the proposed CNN-based DeepGaitV2 series and Transformer-based SwinGait series exhibit significant performance improvements on Gait3D and GREW. 
As for the constrained gait datasets, the DeepGaitV2 series also reaches a new state-of-the-art in most cases, convincingly showing its practicality and generality. 
The source code is available at https://github.com/ShiqiYu/OpenGait.
\end{abstract}

\section{Introduction}
Gait is a biometric characteristic that presents the unique walking patterns of individuals.
Compared with other biometric modalities, 
\textit{e.g.}, face, fingerprint, and iris, 
gait is hard to disguise and can be easily captured at a distance in non-intrusive ways.
For these advantages, 
gait recognition is one of the most feasible techniques for security applications, 
such as crime investigation, suspect tracking, and identity verification~\cite{nixon2006automatic}.

\begin{figure}[tb]
\centering
\includegraphics[width=\linewidth]{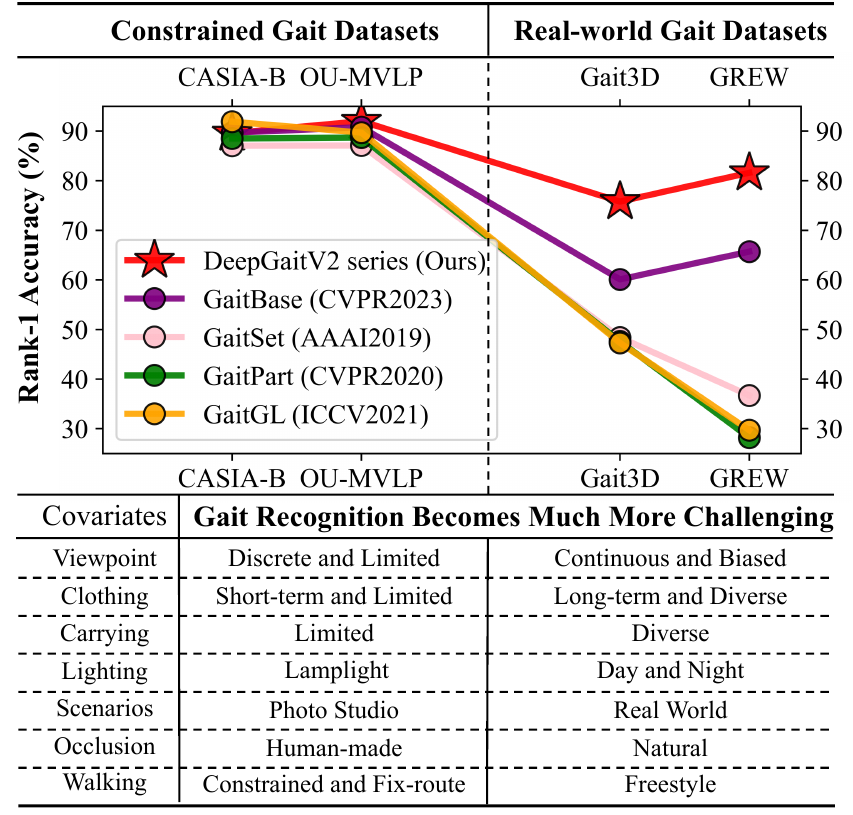}
\caption{
Constrained \textit{v.s.} real-world gait recognition.
}
\label{fig.intro}
\end{figure}
Similar to many vision technologies~\cite{ren2015faster, long2015fully, wang2016temporal, cao2017realtime}, the widely-verified effectiveness and robustness of deep architectures have significantly advanced gait recognition~\cite{wu2016comprehensive, chao2019gaitset}.
However, it is observed that the performance of existing state-of-the-art gait methods undesirably drops dramatically when shifting from previous constrained testing in the lab~\cite{yu2006framework, takemura2018multi} to nowadays real applications in-the-wild~\cite{zhu2021gait, zheng2022gait}, as illustrated in Figure~\ref{fig.intro}. 
Intuitively, this phenomenon should be mainly caused by emerging real-world complexities, which make gait recognition much more challenging than that depicted by prior constrained gait datasets. 
More importantly, our further investigations demonstrate that several underlying stereotypes in gait modeling seriously block the potential of deep gait models. 

First of all, gait data is typically presented as sparse input, such as binary images with a size of $64\times44$, which might create the impression that gait recognition is an `easy task' that shallow networks can handle, as researchers used to do. 
In this paper, we agree that the gait recognition defined by constrained gait datasets, such as the CASIA-B, OU-MVLP, and the latest SUSTech1K~\cite{Shen_2023_CVPR} and CCPG~\cite{Li_2023_CVPR} datasets, can indeed be considered as such `easy' cases that shallow gait models can deal with.
This is because they merely contain several limited covariates manipulated by researchers, whereas gait models can circumvent artificially created challenges by focusing on evidently unchanged characteristics rather than subtle walking patterns. 
However, the scenario takes a significant turn when we consider emerging real-world gait applications.
As depicted in Figure~\ref{fig.intro}, it reveals that the real gait data distribution is comprehensive, diverse, and biased, making the recognition task much more challenging than ever before. 
Results show that shallow gait models are no longer competent for these situations. 
Therefore, the primary goal of this paper is to explore new empirical principles for deep gait model designs. 
To this end, we present a comprehensive study showing that gait models are \textit{depth-hungry}, as most vision tasks desired but prior gait works have been ignored. 
As a result, the proposed \textit{DeepGaitV2} series outperforms previous shallow gait models by a breakthrough margin, showing promising results and relighting the potential of gait recognition for real-world applications.

\begin{figure}[tb]
\centering
\includegraphics[width=0.9\linewidth]{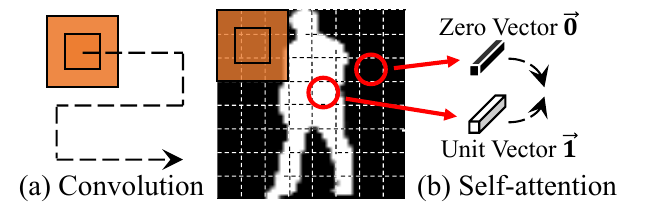}
\caption{
Dumb patch issue on the gait silhouette.
}
\label{fig.dumb}
\end{figure}

Another crucial aspect of gait recognition is learning temporal changes that capture the walking patterns of individuals, especially when appearance features become relatively unreliable, such as cloth-changing and complex occlusion cases.
Currently, two methodologies prevail in the gait research community, \textit{i.e.}, regarding the gait as a set~\cite{chao2019gaitset} or as a sequence. These two kinds of works have performed similar achievements in the past few years by using relatively shallow nets. Among them, the set-based methods attract relatively more attention thanks to their advantages of flexible input and concise architecture. 
But in this paper, we experimentally confirm the superiority of exact characterizing dependencies between successive gait frames, showing that gait models are \textit{temporal-information-hungry}. 

Furthermore, with a modeling shift from CNNs to Transformers~\cite{vaswani2017attention, dosovitskiy2020image, liu2021swin}, the potential of deep gait transformers in addressing practical complexities becomes an attractive prospect. 
However, despite this trend, the gait community has yet to produce influential results verifying the effectiveness of transformers in gait description. 
In this paper, we argue that the neglect of `dumb' patches, illustrated in Figure~\ref{fig.dumb}, lies in the central issue. 
Specifically, unlike the float-encoding RGB image, most areas of the gait silhouette consist of all-white and all-black regions, leading to massive non-informative (dumb) patches. 
CNNs can circumvent this challenge by their natures of dense feature extraction and receptive field expansion through the sliding-window mechanism and hierarchical architecture, respectively. 
But for typical Transformers, the self-attention computation quadratically increases the risk of generating useless or even invalid gradients, as these corrupted cases will occur whenever any single patch within the query-key pair belongs to the dumb patches. 
To address this critical concern, we provide a straight strategy and thus build \textit{SwinGait}.
To our knowledge, SwinGait is the first transformer-based gait method that surpasses previous CNN-based gait methods by a significant margin on various large-scale real-world gait datasets. 
Also, SwinGait outperforms our DeepGaitV2 series in many cases, convincingly exhibiting the promising future of deep gait transformers in outdoor applications. 

Overall, this paper explores new empirical principles for deep gait model designs, \textit{i.e.}, challenges the stereotype of shallow gait models, and demonstrates the superiority of explicit temporal modeling and deep gait transformers. 
Notably, some features of this work warrant highlighting: 
\begin{itemize}
    \item \textbf{Emphasis on generality}: To ensure the applicability of our conclusions, we have chosen to employ the widely accepted block and gait framework, rather than relying on specially designed ones. 
    This decision can enhance the practicability and commonality of our contributions.
    \item \textbf{Insightful discoveries and practical solutions}: This paper stands out through its focus on insightful discoveries and pragmatic solutions, rather than fancy network designs. 
    By addressing key challenges straightforwardly, DeepGaitV2 and SwinGait achieve significant advancements in gait recognition. 
    We draw inspiration from some impressive works that seem simple but significantly promote follow-up studies~\cite{szegedy2016rethinking}.
\end{itemize}

\section{Related Works}

According to the classical taxonomy, gait recognition methods can be broadly divided into model-based and appearance-based categories.
The former typically takes the estimated underlying structure of the human body, such as 2D/3D pose~\cite{liao2020model} and SMPL model~\cite{li2020end}, as input, while the latter tends to extract the gait features from silhouettes directly.
We concentrate on the latter one thanks to its relatively superior performance. 
The brief literature review is carried out from aspects of deep spatial, temporal, and transformer approaches for gait description.

\begin{figure}[t]
\centering
\includegraphics[width=\linewidth]{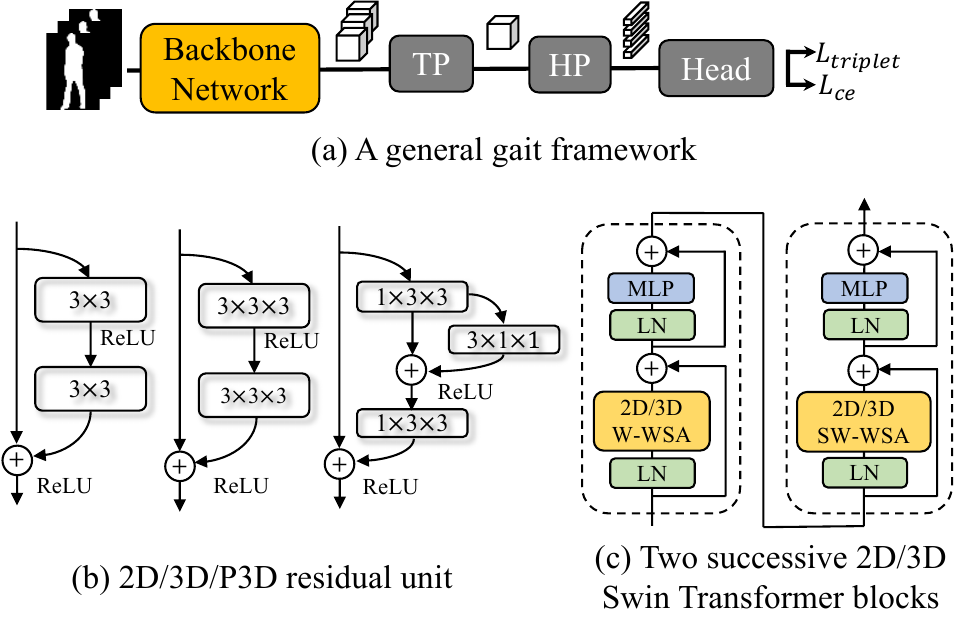}
\caption{The gait framework and basic blocks employed for building DeepGaitV2 and SwinGait series. }
\label{fig.employed} 
\end{figure}

\noindent \textbf{Deep spatial feature extraction.}
Wu~\textit{et al.}~\cite{wu2016comprehensive} were among the pioneering researchers to introduce deep networks for gait recognition.
Thereafter, Chao~\textit{et al.}~\cite{chao2019gaitset} adopted a part-based model for human body description, which is still widely used today.
Fan~\textit{et al.}~\cite{fan2020gaitpart} suggested that gait models should focus on more local details and present the FConv layer.
Lin~\textit{et al.}~\cite{lin2021gait} highlighted the importance of combining spatially local and global features for discriminative gait pattern extraction.
Fan~\textit{et al.}~\cite{Fan_2023_CVPR} deeply reviewed some of the above methods and further proposed a simple yet strong baseline model termed GaitBase.
Despite achieving notable success, these representative methods have not produced competitive performances on the emerging real-world gait datasets, in contrast to the achievements seen on previous constrained gait datasets. 
This paper considers that the root of this problem lies in the stereotypical reliance on shallow gait models. 

\noindent \textbf{Deep temporal modeling} is always one of the most attractive topics for gait recognition because of the specification of walking movement, \textit{i.e.}, periodicity.
Two popular methodologies exist in the literature: set-based~\cite{chao2019gaitset} and sequence-based methods.
The former believes the order information of the gait video is not necessary for presenting the unique individual characteristics since people can roughly identify the position of each frame in the gait cycle according to the appearance.
Practically, the set-based methods extract silhouette features frame by frame and then learn a sequence-level feature using a max-pooling function along the temporal dimension. 
Although the sequence-based methods that explicitly capture features from consecutive frames generally achieve slightly better performance~\cite{fan2020gaitpart, lin2021gait, dou2022metagait}, 
the set-based methods are preferred due to their superior efficiency and robustness to noise. 
In this paper, we deeply discuss these two kinds of methods and experimentally emphasize the superiority of explicit temporal modeling.

\noindent \textbf{Deep gait transformer.}
Several works attempt to introduce the deep transformer into gait recognition, \textit{e.g.}, 
Mogan \textit{et al.}~\cite{mogan2022gait}, and Pinvcic \textit{et al.}~\cite{pinvcic2022gait} directly employ the prevailing ViT model~\cite{dosovitskiy2020image} on gait silhouettes, and Cui \textit{et al.}~\cite{cui2022} use sequence transformer for gait temporal modeling.
Currently, the transformer-based methods did not achieve impressive results on the popular testing benchmarks, especially for the challenging outdoor ones. 

\section{Deep Models for Practical Gait Recognition}
\label{sec.better_deep_model}
In-the-wild settings present a significantly more challenging environment for gait recognition compared to previous controlled settings, rendering existing state-of-the-art methods inferior for real-world applications.
It is now of utmost importance to break prior stereotypes that seriously block the potential of gait models. 
To this end, this paper presents a comprehensive study to explore new empirical principles for deep gait designs from a systematic standpoint. 

As shown in Figure~\ref{fig.employed} (a), we employ a general gait framework summarized by OpenGait~\cite{Fan_2023_CVPR}.
To make it clear and brief, this paper focuses on the designs of the backbone network while keeping other parts unchanged. 
Figure~\ref{fig.employed} (b) presents the 2D, 3D, and pseudo 3D residual units used for constructing deep gait models.
The pseudo 3D residual unit is composed of a 1D temporal and two 2D spatial convolution layers. 
Furthermore, the popular Swin Transformer~\cite{liu2021swin} and its 3D variant shown in Figure~\ref{fig.employed} (c) are also incorporated to construct deep gait transformers. 

\begin{table}[t]
\centering
\caption{
Architectures of DeepGaitV2-3D series. 
Down-sampling is performed by Stage2 and Stage3 with a stride of 2. 
$T$ and $C$ respectively denote the input sequence length and arbitrary channel number.
}
\resizebox{0.51\textwidth}{28mm}{
\begin{tabular}{c|c|ccccc}
\toprule
\multirow{2}{*}{Layers}                                 & \multirow{2}{*}{\begin{tabular}[c]{@{}c@{}}Output feature \\ map size\end{tabular}} & \multicolumn{5}{c}{DeepGaitV2-3D} \\ \cmidrule{3-7} 
& & \multicolumn{1}{c|}{Block structure} & \multicolumn{1}{c|}{10-layer} & \multicolumn{1}{c|}{14-layer} & \multicolumn{1}{c|}{22-layer} & 30-layer \\ \midrule 
Conv0                                                  & ($T$, $C$, 64, 44)               & \multicolumn{5}{c}{$3\times3$, stride 1} \\ \midrule 
Stage1                                                 & ($T$, $C$, 64, 44)               & \multicolumn{1}{c|}{$\begin{bmatrix}
3\times3, C\\ 
3\times3, C
\end{bmatrix}$}      & \multicolumn{1}{c|}{$\times1$}         & \multicolumn{1}{c|}{$\times1$}         & \multicolumn{1}{c|}{$\times1$}         & $\times1$         \\ \midrule 
Stage2                                                 & ($T$, $2C$, 32, 22)              & \multicolumn{1}{c|}{$\begin{bmatrix}
3\times3\times3, 2C\\ 
3\times3\times3, 2C
\end{bmatrix}$}      & \multicolumn{1}{c|}{$\times1$}         & \multicolumn{1}{c|}{$\times2$}         & \multicolumn{1}{c|}{$\times4$}         &  $\times4$        \\ \midrule 
Stage3                                                 & ($T$, $4C$, 16, 11)              & \multicolumn{1}{c|}{$\begin{bmatrix}
3\times3\times3, 4C\\ 
3\times3\times3, 4C
\end{bmatrix}$}      & \multicolumn{1}{c|}{$\times1$}         & \multicolumn{1}{c|}{$\times2$}         & \multicolumn{1}{c|}{$\times4$}         & $\times8$         \\ 
\midrule 
Stage4                                                 & ($T$, $8C$, 16, 11)              & \multicolumn{1}{c|}{$\begin{bmatrix}
3\times3\times3, 8C\\ 
3\times3\times3, 8C
\end{bmatrix}$}      & \multicolumn{1}{c|}{$\times1$}         & \multicolumn{1}{c|}{$\times1$}         & \multicolumn{1}{c|}{$\times1$}         & $\times1$         \\  \midrule 
TP                                                     & (1, $8C$, 16, 11)               & \multicolumn{5}{c}{Temporal Pooling}                                                                                                                                                                                                                                                                             \\ \midrule 
HP                                                     & (1, $8C$, 16, 1)                & \multicolumn{5}{c}{Horizontal Pooling}                                                                                                                                                                                                                                                                             \\ \midrule 
\begin{tabular}[c]{@{}c@{}}Head \end{tabular} & (1, $8C$, 16, 1)                & \multicolumn{5}{c}{Flatten, 16 separate fully-connected layers and BNNecks}                                                                                                                                                                                                                                                                              \\ \bottomrule 
\end{tabular}
}
\label{tab.deepgait}
\end{table}

\begin{figure}[t]
\centering
\includegraphics[width=\linewidth]{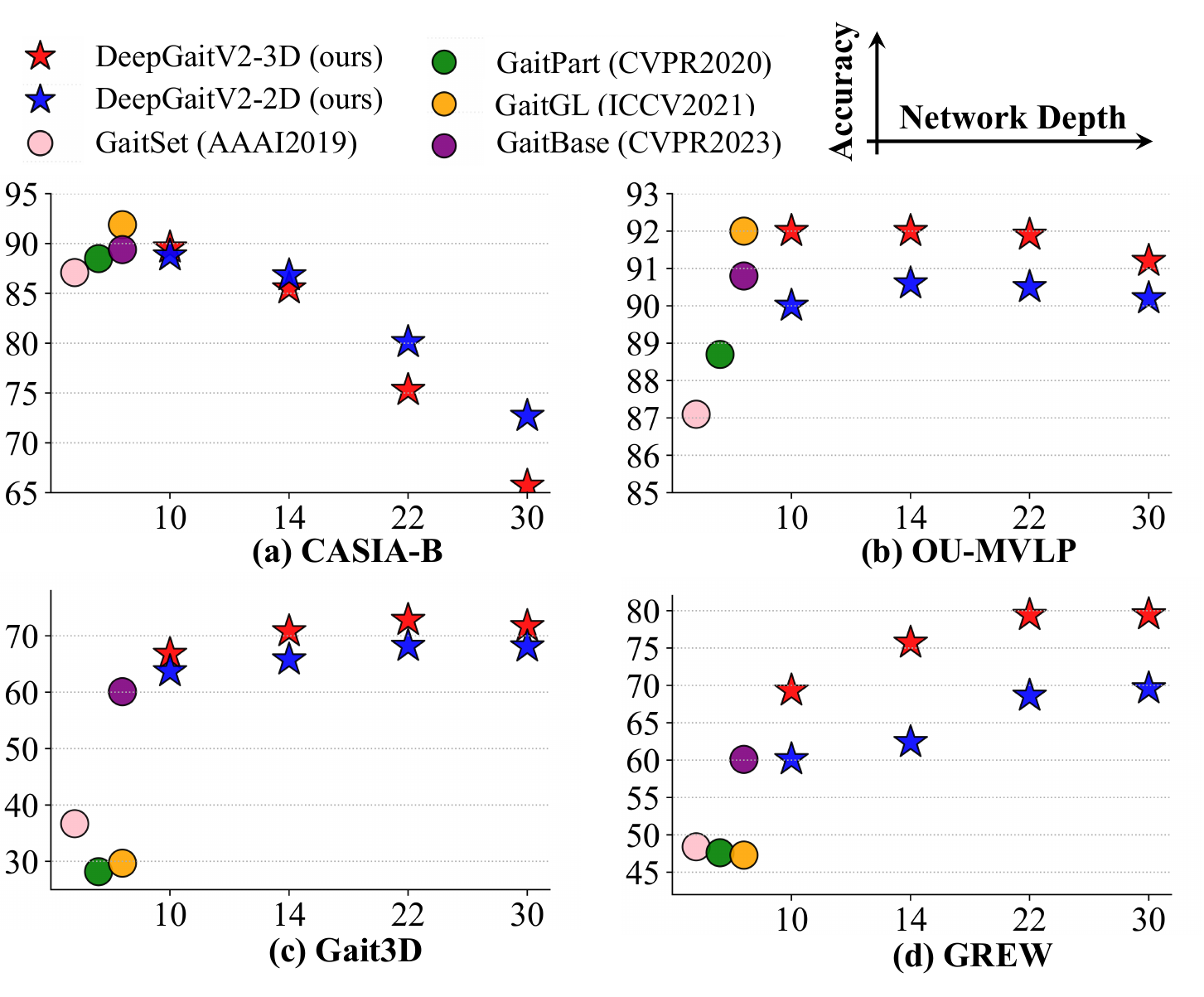}
\caption{
Rank-1 accuracy of DeepGaitV2-2D and 3D with their backbone network going deeper.
The performance of several state-of-the-art methods is introduced for reference. 
}
\label{fig.results} 
\end{figure}

\subsection{Breaking the Stereotype of Shallow Gait Models}
\label{sec.capacity}

As shown in Table~\ref{tab.deepgait}, we build a DeepGaitV2-3D series\footnote{We regard GaitBase in \cite{Fan_2023_CVPR} as a pioneering attempt to explore deep ResNet for gait modeling and call it DeepGaitV1. 
Hence, our CNN-based gait models are termed DeepGaitV2.} by stacking five layers, \textit{i.e.}, the initial convolution (Conv0) and the following four stages (Stage 1 to 4). 
The number of residual blocks stacked on each stage, denoted as $B$, determines the network depth.
For instance, if $B$ is set to [1,1,1,1], the network depth should be 2$\times$($1$+$1$+$1$+$1$)+2=10, where the last +2 implies the initial Conv0 and final head layers.
As a result, Table~\ref{tab.deepgait} presents a serial of DeepGaitV2-3D with a depth ranging from 10 to 30.
Note that Stage 1 of DeepGaitV2-3D is formed by 2D residual units for computational efficiency. 
DeepGaitV2-2D is made by replacing all the 3D blocks with their 2D counterparts.

\begin{figure}[tb]
\centering
\includegraphics[width=\linewidth]{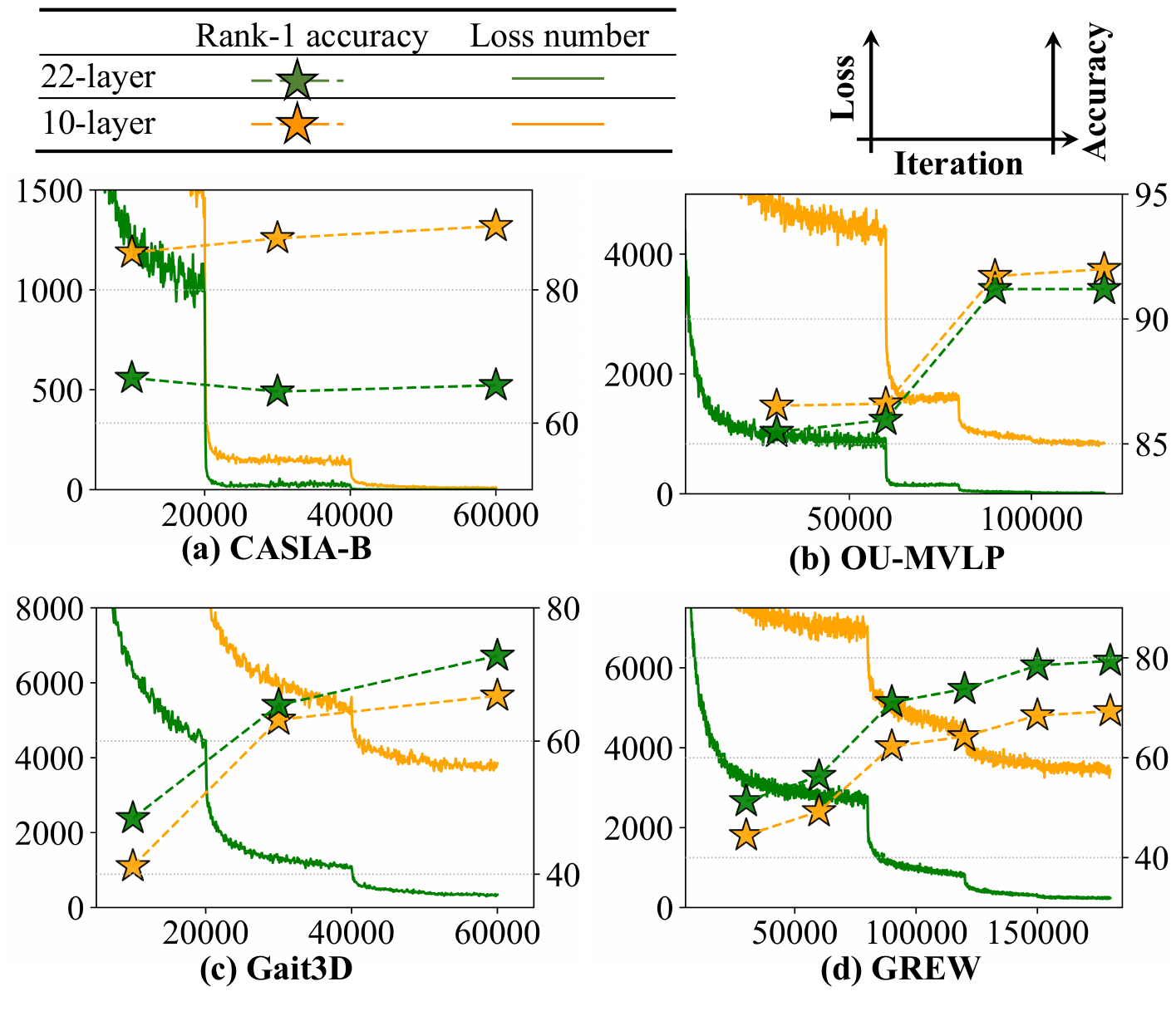}
\caption{
The DeepGaitV2-3D series meets the over-fitting cases on (a) CASIA-B and (b) OU-MVLP, with the network depth increasing. 
The loss number presents the count of triplets that cause non-zero loss in the training batch, directly reflecting the network's convergence state.
}
\label{fig.overfitting} 
\end{figure}

As illustrated in Figure~\ref{fig.results}, there exists a noticeable distinction in performance trends between the constrained and real-world gait evaluations. 
Specifically, applications on Gait3D and GREW exhibit substantial benefits from deep models, while contrasting outcomes are observed on CASIA-B and OU-MVLP.
The further analysis shown in Figure~\ref{fig.overfitting} expose the over-fitting cases on CASIA-B and OU-MVLP, \textit{i.e.}, the 22-layer DeepGaitV2-3D converges better but performs worse than its 10-layer counterpart. 

Moreover, we have found similar over-fitting phenomena on the latest constrained gait datasets proposed at CVPR2023, namely SUSTech1K and CCPG.
Despite the introduction of new human-controlled challenges, such as night illumination, complex occlusion, and diverse clothing changes, deep models continue to exhibit over-fitting behavior on these gait datasets. 

Though the above successes were achieved by shallow models on constrained gait datasets, we argue that this stereotype seriously hinders the practical applications of gait recognition. 
Constrained gait datasets typically have a limited number of covariates controlled by researchers, which inevitably simplifies the gait recognition task. 
While this data collection strategy has been instrumental in early gait research, enabling the exploration of key factors influencing recognition accuracy through controllable annotation and data distribution, it also results in a drawback. 
Gait models might overlook subtle walking patterns and instead focus on capturing obvious, unchanged, gait-unrelated characteristics.
For instance, CCPG~\cite{Li_2023_CVPR} demonstrated that deep networks tend to focus on faces and shoes when diverse clothing-changing covariates are introduced. 
Consequently, gait recognition used to be perceived as an `easy' task for shallow models to tackle. 

However, in the case of emerging real-world gait datasets, their comprehensive, diverse, and biased distribution significantly amplifies the challenges of gait recognition. 
As evidenced by Figure~\ref{fig.results} (c) and (d), a clear need for deep networks becomes evident. 
Therefore, this paper suggests that forthcoming research should embrace deep models as the foundational architecture for real-world gait applications.
To this end, we recommend utilizing the 10-layer and 22-layer DeepGaitV2 models as baseline benchmarks for constrained and real-world evaluations, respectively.

\subsection{Superiority of Explicit Temporal Modeling}
\label{sec.temporal}
Capturing body changes during the walking process is critical for gait description.
In recent literature, a popular methodology involves treating gait as an unordered set of silhouettes~\cite{chao2019gaitset}.
Its core idea comes from the observation that people can easily identify a certain temporal position of a silhouette in the gait cycle merely by looking at its appearance. 
This is possible because the walking process is typically periodic. 
As a result, the set-based methods consider order information unnecessary and tend not to explicitly utilize motion features among neighboring frames. 
On the other hand, the sequence-based methods usually pay much attention on designing temporal models. 

\begin{figure}[tb]
\centering
\includegraphics[width=0.9\linewidth]{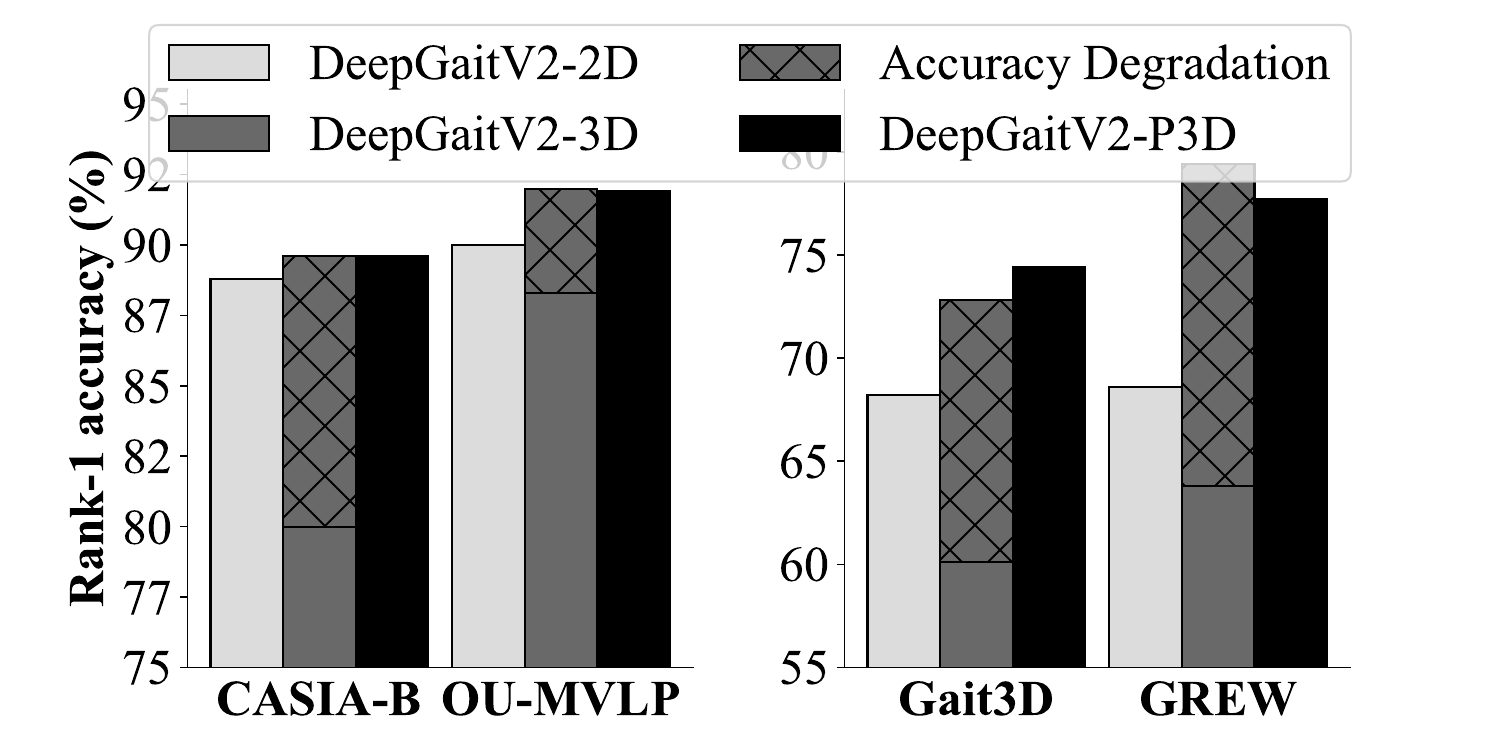}
\caption{
Rank-1 accuracy of DeepGaitV2-2D/3D/P3D. The shaded area presents the accuracy degradation of DeepGaitV2-3D when shuffling input frames.
}
\label{fig.temporal} 
\end{figure}

In this paper, DeepGaitV2-2D and 3D respectively present the plain set-based and sequence-based DeepGaitV2, where the former only consists of 2D residual blocks, and the latter is mostly composed by 3D blocks. 
Figure~\ref{fig.results} shows the sequence-based DeepGaitV2 clearly surpasses its set-based counterpart on various mainstream gait datasets. 
To further confirm the effectiveness of explicit temporal modeling, we conduct the following experiments: 
\begin{itemize}
    \item Upon shuffling input frames, thus disrupting the dependencies between adjacent silhouettes, a noticeable degradation in accuracy becomes evident. This is illustrated by the shaded region in Figure~\ref{fig.temporal}, wherein the accuracy drops are noteworthy: -9.6\%, -3.7\%, -12.7\%, and -15.8\% for CASIA-B, OU-MVLP, Gait3D, and GREW, respectively. These results compellingly show the significance of sequential characteristics for gait modeling.
    \item To exclude the potential benefits brought by increasing parameters and computational costs of 3D blocks, we develop a lightweight counterpart of DeepGaitV2 termed P3D\footnote{
    \# param: 27.5 \textit{v.s.} 9.3 \textit{v.s.} 11.1 MB, 
    and FLOPs: 6.8 \textit{v.s.} 2.4 \textit{v.s.} 2.9 GFLOPs, per silhouette image within DeepGaitV2-3D \textit{v.s.} 2D \textit{v.s.} P3D. 
    Here we only consider the backbone. The same below.}.
    As shown in Figure~\ref{fig.temporal}, DeepGaitV2-P3D achieves a competitive or even superior performance than its 3D counterpart.
    Moreover, it outperforms its 2D counterpart obviously, \textit{i.e.}, +0.8\%, +1.9\%, +6.2\%, +9.1\%, on CASIA-B, OU-MVLP, Gait3D and GREW, respectively. 
\end{itemize}

In summary, this paper regards the order of gait frames as the inherent time stamps, precisely guiding the fusion of locally adjacent frames for fine-grained extraction of motion features. 
The results emphasize the deals of adequately using sequential priors instead of completely ignoring them. 
In our experiments, DeepGaitV2-3D and its P3D counterpart serve as two direct examples of sequence-based gait models, not only showcasing their effectiveness but also endowing the conclusions with generality and practicality.

\begin{table*}[tb]
\centering
\caption{
Implementation details. \#LR and \#WD present the initial learning rate and weight decay. $T$ and $C$ respectively denote the input sequence length and arbitrary channel number shown in Table~\ref{tab.deepgait} and Figure~\ref{fig.swingait}. The batch size ($q, k$) indicates $q$ subjects with $k$ sequences per subject. The $I_{\textrm{max}}$ presents the maximum number of iterations per half-cosine period.
}
\resizebox{1.0\textwidth}{!}{
\begin{tabular}{c|c|cccc|cccc}
\toprule
\multirow{4}{*}{Datasets} & \multirow{4}{*}{Batch size} & \multicolumn{4}{c|}{DeepGaitV2 series, $T$=30, $C$=64}                                                                                                                                                                 & \multicolumn{4}{c}{SwinGait series, $T$=30, $C$=64}                                                                                                                                                                                                                         \\ \cmidrule{3-10} 
                          &                             & \multicolumn{1}{c|}{\multirow{2}{*}{Optimizer}}                                               & \multicolumn{3}{c|}{Scheduler}                                                                         & \multicolumn{1}{c|}{\multirow{2}{*}{Optimizer}}                                                 & \multicolumn{3}{c}{Scheduler}                                                                                                                               \\ \cmidrule{4-6}  \cmidrule{8-10} 
                          &                             & \multicolumn{1}{c|}{}                                                                         & Type                                                                   & Milestones      & Total steps & \multicolumn{1}{c|}{}                                                                           & Type                                                                                       & $I_{\textrm{max}}$ & \begin{tabular}[c]{@{}c@{}}Total steps\end{tabular} \\ \midrule 
CASIA-B                   & (8, 16)                        & \multicolumn{1}{c|}{\multirow{6}{*}{\begin{tabular}[c]{@{}c@{}}SGD\\ \#LR=$1\times10^{-1}$ \\ \#WD=$5\times10^{-5}$ \end{tabular}}} & \multirow{6}{*}{\begin{tabular}[c]{@{}c@{}}Multi-step \\ (drop by $1/10$\\ per milestone)\end{tabular}} & [20k, 40k, 50k]   & 60k         & \multicolumn{1}{c|}{\multirow{6}{*}{\begin{tabular}[c]{@{}c@{}}AdamW \\ \#LR=$3\times10^{-4}$ \\ \#WD=$2\times10^{-2}$ \\ \#LR$_{min}$=$3\times10^{-5}$  \end{tabular}}} & \multirow{6}{*}{\begin{tabular}[c]{@{}c@{}}Cosine-\\ annealing\\ (update per\\ 1k steps) \end{tabular}} & -      & -                                                     \\
OU-MVLP                   & (32, 8)                        & \multicolumn{1}{c|}{}                                                                         &                                                                        & [60k, 80k, 100k]  & 120k        & \multicolumn{1}{c|}{}                                                                           &                                                                                            & -      & -                                                     \\
CCPG                   & (8, 16)                        & \multicolumn{1}{c|}{}                                                                         &                                                                        & [20k, 40k, 50k]  & 60k        & \multicolumn{1}{c|}{}                                                                           &                                                                                            & 60k      & 80k                                                     \\
SUSTech1K                   & (8, 8)                        & \multicolumn{1}{c|}{}                                                                         &                                                                        & [20k, 30k, 40k]  & 50k        & \multicolumn{1}{c|}{}                                                                           &                                                                                            & 50k      & 60k                                                     \\
Gait3D                    & (32, 4)                        & \multicolumn{1}{c|}{}                                                                         &                                                                        & [20k, 40k, 50k]   & 60k         & \multicolumn{1}{c|}{}                                                                           &                                                                                            & 60k    & 80k                                                   \\
GREW                      & (32, 4)                        & \multicolumn{1}{c|}{}                                                                         &                                                                        & [80k, 120k, 150k] & 180k        & \multicolumn{1}{c|}{}                                                                           &                                                                                            & 150k   & 200k                                                  \\ \bottomrule 
\end{tabular}
}
\label{tab.implementation}
\end{table*}

\subsection{Superiority of Deep Gait Transformer}
While CNN-based architectures continue to dominate gait recognition, 
the potential of deep gait transformers in handling practical complexities has become an attractive prospect. 
Despite this promising trend, impactful results to verify the effectiveness of transformers in gait description are still lacking within the gait community.

\begin{figure}[t]
\centering
\includegraphics[width=0.8\linewidth]{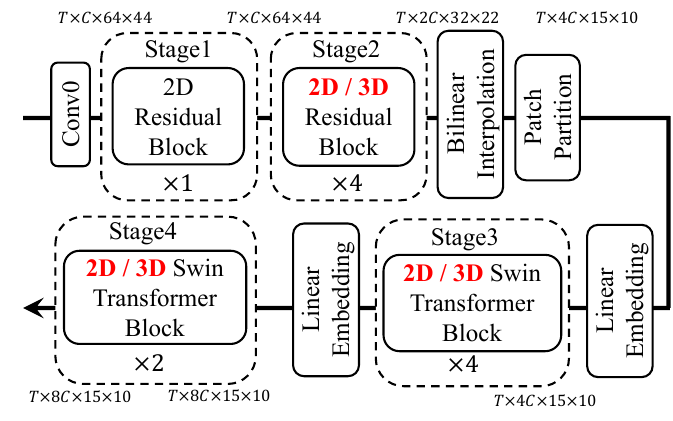}
\caption{Overall architecture of SwinGait-2D/3D.
}
\label{fig.swingait} 
\end{figure}

Deep transformer models typically divide an input RGB image into non-overlapping patches.
Each patch is treated as a `token', and its feature is represented as a concatenation of the raw RGB pixel values. 
The self-attention mechanism or its variant is then performed on the linear embedding of these tokens.
However, for gait recognition, a substantial number of patches generated by directly partitioning the binary silhouette consist of either all-white or all-black regions.
These patches do not provide any edge or shape features and are therefore termed by `dumb' patches. 
Further statistical analysis reveals that 84.1\% of patches in the GREW dataset are this kind of useless patches when using an image size of $64 \times 64$ and a patch size of $4 \times 4$. 
Since all values within a dumb patch are either zero or one, these patches render backward gradients characteristically useless or even computationally invalid for optimizing the training parameters of bottom transformer layers. 
To address this issue, three straightforward solutions have been considered: 
\begin{itemize}
    \item Using a large patch size. 
    By setting patch size to $16 \times 16$, the proportion of dumb patches in the GREW dataset decreases to 46.2\%. 
    Nevertheless, this means the silhouette will be directly down-sampled by 1/16 at the input layer. 
    In this way, only $\frac{64}{16}\times\frac{64}{16}=16$ tokens will be fed into the following transformer backbone, making it difficult to hierarchically capture the fine-grained gait features. 
    \item Copying the convolution mechanism, \textit{i.e.}, partitioning the silhouette by a sliding window with a small stride. 
    This strategy can \textit{linearly} reduce the proportion of dumb patches. 
    However, the risk of generating useless or even invalid gradients still increases \textit{quadratically}, as these corrupted cases will occur whenever any single patch within the query-key pair belongs to the dumb patches. 
    \item We adopt another coping strategy shown in Figure~\ref{fig.swingait}, \textit{i.e.}, instantiating the first two stages as convolution blocks.
    Thanks to the sliding-window mechanism and nonlinear modeling capability of the hierarchical convolution layers, Conv0, Stage 1 and Stage 2 can transform the binary silhouette data to the float-encoding feature map to provide dense low-level structural features. 
\end{itemize}

The following transformer layers, \textit{i.e.}, Stage 3 and Stage 4, will then leverage the advanced transformer-based modules to learn the high-level identification features.
We employ the widely-accepted 2D and 3D Swin Transformer blocks~\cite{liu2021swin, liu2022video} shown in Figure~\ref{fig.employed} (c) to construct SwinGait-2D and its 3D counterpart, respectively.
The main reasons are two-fold: 
\begin{itemize}
    \item Performing self-attention in the local spatiotemporal window enables the exploration of local appearance and motion details for fine-grained gait description.
    \item The excellent speed-accuracy trade-off of Swin Transformer attracts us a lot.
\end{itemize}

As shown in Figure~\ref{fig.swingait}, the structures of Conv0, Stage 1 and Stage 2 in SwinGait-2D/3D inherit from that of DeepGaitV2-2D/3D for fair comparisons~\footnote{On the Gait3D dataset, SwinGait-3D inherits DeepGaitV2-P3D since the latter performs better than its 3D counterpart.}.
After these convolution blocks, the feature maps are with the size of $T\times 2C \times 32 \times 22$, where $T$ is the number of frames, and $C$ is the number of channels. 
To match the size limit required by the downstream transformer blocks, these feature maps are resized to $2C \times 30 \times20$ frame by frame via a bilinear interpolation layer.
Then, we treat each 3D patch of size $2C\times2\times2$ as a token.
Thus, the 3D patch partitioning layer obtains $T\times\frac{30}{2}\times\frac{20}{2}$ 3D tokens, with each patch/token consisting of a $8C$-dimensional feature. 
A linear embedding layer is then applied to project each token to a $4C$-dimensional vector.

\begin{figure}[tb]
\centering
\includegraphics[width=\linewidth]{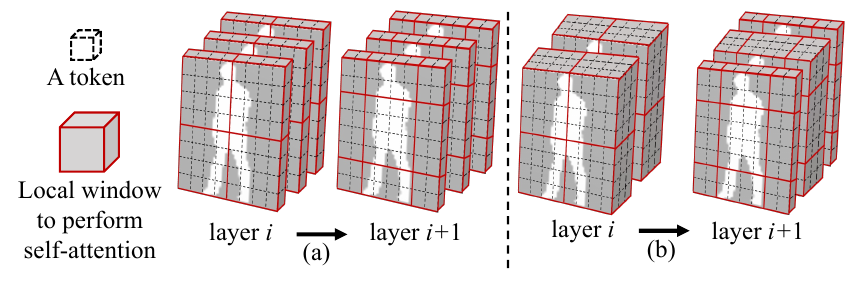}
\caption{
An example of (a) 2D and (b) 3D shifted window with a window size of (a) ($1,4,4$) and (b) ($3,4,4$).
}
\label{fig.window} 
\end{figure}

\begin{table}[tb]
\centering
\caption{
Datasets in use. \#ID and \#Seq present the number of identities and sequences.
}
\resizebox{0.46\textwidth}{!}{
\begin{tabular}{cccccc}
\hline
\multirow{2}{*}{DataSet} & \multicolumn{2}{c}{Train Set} & \multicolumn{2}{c}{Test Set} & \multirow{2}{*}{\begin{tabular}[c]{@{}c@{}}Collection\\ situations\end{tabular}} \\
                         & Id            & Seq           & Id           & Seq           &                                                                                 \\ \hline
CASIA-B                  & 74            & 8,140         & 50           & 5,500         & Constrained                                                            \\
OU-MVLP                  & 5,153         & 144,284       & 5,154        & 144,412       & Constrained                                                                      \\
CCPG                  & 100            & 8,187         & 100           & 8,095         & Constrained                                                              \\
SUSTech1K                  & 200         & 5,988       & 850        & 19,228       & Constrained                                                                      \\
Gait3D                   & 3,000         & 18,940        & 1,000        & 6,369         & Real-world                                                            \\
GREW                     & 20,000        & 102,887       & 6,000        & 24,000        & Real-world                                                            \\ \hline
\end{tabular}
}
\label{tab.datasets}
\end{table}

\begin{table*}[tb]
\centering
\caption{
Comparison to state-of-the-art on CASIA-B, OU-MVLP, Gait3D and GREW. On the CASIA-B and OU-MVLP, the identical-view cases are excluded. The accuracy on CASIA-B is the average of three testing subsets.
}
\scalebox{0.82}{%
    \setlength{\tabcolsep}{0.5em}
\begin{tabular}{ccc|cccccccc}
\toprule
\multirow{4}{*}{Method} & \multicolumn{2}{c}{\multirow{4}{*}{Publication}} & \multicolumn{8}{|c}{Testing Datasets}  \\ \cmidrule{4-11} 
& \multicolumn{2}{c|}{}                        & CASIA-B & \multicolumn{1}{c|}{OU-MVLP} & \multicolumn{3}{c|}{Gait3D}             & \multicolumn{3}{c}{GREW} \\ \cmidrule{4-11} 
                        & \multicolumn{2}{c|}{}                        & R-1     & \multicolumn{1}{c|}{R-1}     & R-1 & R-5 & \multicolumn{1}{c|}{mAP}  & R-1    & R-5   & R-10  \\ \midrule 
GaitSet                 & \multicolumn{2}{c|}{AAAI 2019}               & 87.1    & \multicolumn{1}{c|}{87.1}    & 36.7 & 58.3 & \multicolumn{1}{c|}{30.0} & 46.3   & 63.6   & 70.3   \\
GaitPart                & \multicolumn{2}{c|}{CVPR 2020}               & 88.5    & \multicolumn{1}{c|}{88.7}    & 28.2 & 47.6 & \multicolumn{1}{c|}{21.6} & 44.0   & 60.7   & 67.3   \\
GaitGL                  & \multicolumn{2}{c|}{ICCV 2021}               & 91.9    & \multicolumn{1}{c|}{89.7}    & 29.7 & 48.5 & \multicolumn{1}{c|}{22.3} & 47.3   & -      & -      \\
GaitBase                & \multicolumn{2}{c|}{CVPR 2023}               & 86.1       & \multicolumn{1}{c|}{90.8}       & 64.6 & - & \multicolumn{1}{c|}{-} & 60.1      & -      & -      \\ \bottomrule \toprule 
Proposed                & \#param.                & FLOPs                & \multicolumn{8}{c}{CNN-based DeepGaitV2 series \textit{v.s.} Transformer-based SwinGait series}                                  \\ \midrule 
DeepGaitV2-2D           & 9.3M                  & 2.4G                 & 88.8    & \multicolumn{1}{c|}{90.0}    & 68.2 & 84.6 & \multicolumn{1}{c|}{60.4} & 68.6   & 82.0   & 86.1   \\
SwinGait-2D           & 10.9M                 & 2.5G                & -        & \multicolumn{1}{c|}{-}        & 69.4     & 84.6     & \multicolumn{1}{c|}{61.6}     & 70.8       & 83.7       & 87.6       \\ \midrule 
DeepGaitV2-3D           & 27.5M                 & 6.8G                 & 89.6    & \multicolumn{1}{c|}{\textbf{92.0}}    & 72.8 & 86.2 & \multicolumn{1}{c|}{63.9} & \textbf{79.4}   & \textbf{88.9}   & 91.4   \\
DeepGaitV2-P3D          & 11.1M                 & 2.9G                 & 89.6    & \multicolumn{1}{c|}{91.9}    & 74.4 & \textbf{88.0} & \multicolumn{1}{c|}{65.8} & 77.7   & 87.9   & 90.6   \\
SwinGait-3D           & 13.1M                 & 4.0G                 & -         & \multicolumn{1}{c|}{-}        & \textbf{75.0} & 86.7     & \multicolumn{1}{c|}{\textbf{67.2}}     & 79.3   &\textbf{88.9}   & \textbf{91.8}   \\ \bottomrule 
\end{tabular}
}
\label{tab.results}
\end{table*}

\begin{table*}[t]
\centering
\caption{Evaluation with different attributes on SUSTech1K.}
    \scalebox{0.83}{%
    \setlength{\tabcolsep}{0.5em}

        \begin{tabular}{cc|cccccccc|c|c}
        
        \toprule
        \multirow{2}{*}{Method}   & \multirow{2}{*}{Publication}     & \multicolumn{8}{c}{Probe Sequence (R-1)}      &  \multicolumn{2}{|c}{Overall}                    \\
          &                                                                    & Normal    & Bag    & Clothing  & Carrying  & Umbrella  & Uniform  & Occlusion & Night   & R-1    & R-5  \\         \midrule
        GaitSet      & AAAI 2019         & 69.10     & 68.25   & 37.44     & 65.01     & 63.08      & 61.00    & 67.19     & 23.04   & 65.04  & 84.76  \\
        GaitPart    & CVPR 2019                                & 62.20     & 62.81   & 33.08     & 59.53     & 57.25      & 54.85    & 57.20     & 21.75   & 59.19  & 80.79  \\
        GaitGL        & ICCV 2021                              & 67.11     & 66.16   & 35.92     & 63.31     & 61.58      & 58.07    & 66.59     & 17.88  &  63.14 & 82.82  \\ 
        GaitBase    & CVPR 2023                             & 81.46     & 77.48   & \textbf{49.60}     & 75.77     & 75.55      & 76.66    & 81.40     & 25.92   & 76.12  & 89.39   \\ \midrule
        SwinGait-3D    & \multirow{2}{*}{\begin{tabular}[c]{@{}c@{}}Ours \end{tabular}}                             & 78.00     & 74.07   & 35.02     & 71.41     & 71.83      & 73.48    & 78.05     & 26.72   & 71.48  & 87.62   \\
        DeepGaitV2-P3D         &                        & \textbf{83.53}     & \textbf{79.51}   & 46.28    & \textbf{76.80}     & \textbf{79.15}       & \textbf{78.46}    & \textbf{81.15}     & \textbf{27.30}   & \textbf{77.38}  & \textbf{90.16} \\ 
        \bottomrule
        \end{tabular}
        }
\label{tab:results_on_SUSTech1K}
\end{table*}

Stage 3 and 4 of SwinGait-2D/3D are composed of the standard 2D/3D Swin Transformer blocks. 
Specifically, the 2D Swin Transformer~\cite{liu2021swin} is built by replacing the standard multi-head self-attention module with a module based on shifted windows. 
As shown in Figure~\ref{fig.window} (a), the 2D shifted windows bridge the windows of the preceding layer, providing connections among them.
The 3D Swin Transformer shown in Figure~\ref{fig.window} (b) extends the scope of local attention computation from only the spatial domain to the spatiotemporal domain.
In our implementation, the 3D window size of 2D and 3D Swin Transformer block is set to ($1,3,5$) and ($3,3,5$). 
Moreover, Stage 3 and 4 are connected by another linear embedding layer responsible for mapping the feature of each token to a $8C$-dimensional vector.

The dumb patch problem impacting the optimization process is considered one of the most fundamental challenges for developing the deep gait transformer. 
SwinGait presents a straightforward solution with improvement breakthroughs.

\section{More Experiments}
For a convincing elaboration, some experiments have been briefly described in the previous section along with exploring deep models for gait recognition. In this section, we will first provide the details of the datasets and the implementation for completeness, and then we will present the comparisons with state-of-the-art methods and the ablation study.
\label{sec.experiments}

\noindent \textbf{Datasets.} Six popular gait datasets are employed, and they are CASIA-B, OU-MVLP, CCPG, SUSTech1K, Gait3D, and GREW. 
They were collected in 2006, 2018, 2023, 2023, 2022, and 2021, respectively. The former four datasets were collected in fully constrained laboratories, and the latter two were from real-world scenarios. 
The key statistical indicators are shown in Table~\ref{tab.datasets}.
Our experiments strictly follow the official evaluation protocols. 

\noindent \textbf{Implementation details.} Implementation details are mostly presented in Table~\ref{tab.implementation}. Unless otherwise specified, the silhouettes are aligned by the normalization method used in \cite{takemura2018multi} and resized to $64\times44$. The input sequence is unordered for implementing the DeepGaitV2-2D and SwinGait-2D while ordered for others. The triplet loss with a margin of 0.2 is used. d) The spatial augmentation strategy outlined by \cite{Fan_2023_CVPR} is adopted. The source code will be available. 
Particularly for the SwinGait series, the bottom convolution and head layers are initialized from the corresponding trained DeepGaitV2 models with the learning rate of $3\times10^{-5}$. Following~\cite{liu2021swin}, a stochastic depth rate of $0.1$ is employed.

\noindent \textbf{Comparisons with state-of-the-arts.}
As shown in Table~\ref{tab.results}, Table~\ref{tab:results_on_SUSTech1K}, and Table~\ref{tab:results_on_CCPG}, the DeepGaitV2 series evidently outperform other state-of-the-art methods on the employed datasets except CASIA-B.
Notably, on real-world gait datasets Gait3D and GREW, the DeepGaitV2 series has achieved remarkable performance improvements. 
Moreover, the DeepGaitV2 series also achieves new state-of-the-arts on some constrained gait datasets like OU-MVLP, SUSTech1K, and CCPG, convincingly declaring that exploring a strong baseline model for gait recognition is still not a done deal regardless of gait data collection scenarios. 

More importantly, Table~\ref{tab.results} demonstrates that the SwinGait series surpasses the corresponding CNN-based DeepGaitV2 model for most outdoor cases, considering a speed-accuracy trade-off.
We also note that SwinGait series performs relatively inferior on the constrained gait datasets like SUSTech1K and CCPG, as shown in Tab.~\ref{tab:results_on_SUSTech1K} and \ref{tab:results_on_CCPG}. 
This observation is consistent with that on other vision tasks, \textit{i.e.}, transformers may achieve obvious improvements on the big challenging datasets. 
Overall, we think the evolution of network architectures is always worth keeping a close eye on, and Transformer is particularly noteworthy and popular for its use of attention to model spatial-temporal dependencies. 

\begin{table}[t]
\centering
\caption{Evaluation with different attributes on CCPG.}
    \scalebox{0.82}{%
    \setlength{\tabcolsep}{0.5em}

        \begin{tabular}{cc|cc|cc|cc}
        
        \toprule
        \multirow{2}{*}{Model}   & \multirow{2}{*}{Publication}     & \multicolumn{2}{c|}{CL-Full}      &  \multicolumn{2}{c|}{CL-UP} &  \multicolumn{2}{c}{CL-DN}  \\
          & & R-1 & mAP  & R-1 & mAP & R-1 & mAP                \\         \midrule[1pt]
          GaitSet        & AAAI 2019 & 77.7 & 46,4 & 83.5 & 59.6 & 83.2 & 61.4 \\ 
          GaitPart       & CVPR 2020 & 77.8 & 45.5 & 84.5 & 63.1 & 83.3 & 60.1 \\ 
          GaitGL         & ICCV 2021 & 69.1 & 27.0 & 75.5 & 37.1 & 77.6 & 37.6 \\ 
          AUG-OGBase     & CVPR 2023 & 84.7 & 52.9 & 88.4 & 67.5 & 89.4 & 67.9 \\ \midrule
          SwinGait3D     & \multirow{2}{*}{\begin{tabular}[c]{@{}c@{}}Ours \end{tabular}} & 80.7 & 49.3 & 92.0 & 72.1 & 83.7 & 67.7 \\
          DeepGaitV2-P3D     &     & \textbf{91.1} & \textbf{63.6} & \textbf{95.3} & \textbf{81.1} & \textbf{92.9} & \textbf{79.4} \\ 
          \bottomrule
        \end{tabular}
        }
\label{tab:results_on_CCPG}
\end{table}

\noindent \textbf{Ablation study}. 
The controlling settings, which directly determine the backbone capacity, have been considered, \textit{i.e.}, the network depth and width.
Here, we further widen the DeepGaitV2 series, \textit{i.e.}, enlarging $C$ from 32 to 128 as shown in Table~\ref{tab.ablation}.
Results show that the accuracy improves steadily, but the model size and computation cost increase dramatically.
To improve the efficiency, we suggest the DeepGaitV2-P3D with $C$=64 as the baseline model for the DeepGaitV2 series for a good balance.
As for the SwinGait series, Table~\ref{tab.ablation} shows that reasonably deepening the backbone can also bring performance gain.

\begin{table}[tb]
\centering
\caption{
Ablation study of the DeepGaitV2 and SwinGait series on Gait3D and GREW. The $C$ and $B$ respectively denote the arbitrary channel number and the block numbers.
}
\resizebox{0.48\textwidth}{!}{
\begin{tabular}{c|l|c|c|l|l}
\toprule
\multirow{2}{*}{Method}         & \multirow{2}{*}{\begin{tabular}[c]{@{}c@{}}Control\\ Condition\end{tabular}} & \multirow{2}{*}{\begin{tabular}[c]{@{}c@{}}Gait3D\\ R-1 \end{tabular}} & \multirow{2}{*}{\begin{tabular}[c]{@{}c@{}}GREW\\ R-1 \end{tabular}} & \multirow{2}{*}{\#param.} & \multirow{2}{*}{FLOPs} \\
                                &                                                                              &                                                                       &                                                                     &                         &                        \\ \midrule 
\multirow{3}{*}{\begin{tabular}[c]{@{}c@{}}DeepGaitV2-3D \end{tabular}}   & $C$=32                                                                         & 69.4                                                                      & 73.1                                                                    & 6.9M                        & 1.7G                       \\
                                & $C$=64                                                                         & 72.8                                                                      & 79.4                                                                    & 27.5M                        & 6.8G                       \\
                                & $C$=128                                                                        & \textbf{75.8}                                                                      & \textbf{81.6}                                                                    & 109.8M                        & 27.2G                        \\ \midrule 
\multirow{3}{*}{\begin{tabular}[c]{@{}c@{}}DeepGaitV2-P3D \end{tabular}}  & $C$=32                                                                         & 67.9                                                                      & 72.8                                                                    & 2.8M                        & 0.7G                       \\
                                & $C$=64                                                                         & 74.4                                                                      & 77.7                                                                    & 11.1M                        & 2.9G                       \\
                                & $C$=128                                                                        & 75.0                                                                      & 81.0                                                                    & 44.4M                        & 11.4G                       \\ \midrule 
\multirow{2}{*}{\begin{tabular}[c]{@{}c@{}}SwinGait-3D \end{tabular}}  & $B$=[1,2,2,2]                                                                      & 69.0                                                                      & 77.3                                                                    & 9.8 M                          & 2.5G                      \\
                                & $B$=[1,4,4,2]                                                                      & 75.0                                                                      & 79.3                 & 13.1M      & 4.0G                                             \\ \bottomrule 
\end{tabular}
}
\label{tab.ablation}
\end{table}

\section{Discussion and Future Work}
Apart from achieving performance breakthroughs on real-world datasets, more importantly, this study summarizes critical empirical principles for deep gait model construction by straightforward solutions, based on extensive experimental analysis and discussion.
With considering practicality and generality, we break the stereotype of shallow gait models, and demonstrate the superiority of explicit temporal modeling and deep gait transformers.
While our achievement of an impressive 81.6\% rank-1 accuracy on the GREW dataset is notable, it is clear that bridging the gap for practical gait recognition remains a significant task.
Additionally, collecting and labeling gait video data is more expensive and difficult than most other vision tasks, making it a major challenge to improve gait recognition in the future. 
Feasible solutions are expected.

\newpage
{\small
\bibliographystyle{ieee_fullname}
\bibliography{egbib}

\begin{thebibliography}{10}\itemsep=-1pt

\bibitem{cao2017realtime}
Zhe Cao, Tomas Simon, Shih-En Wei, and Yaser Sheikh.
\newblock Realtime multi-person 2d pose estimation using part affinity fields.
\newblock In {\em Proceedings of the IEEE conference on computer vision and pattern recognition}, pages 7291--7299, 2017.

\bibitem{chao2019gaitset}
Hanqing Chao, Yiwei He, Junping Zhang, and Jianfeng Feng.
\newblock Gaitset: Regarding gait as a set for cross-view gait recognition.
\newblock In {\em Proceedings of the AAAI conference on artificial intelligence}, volume~33, pages 8126--8133, 2019.

\bibitem{cui2022}
Yufeng Cui and Yimei Kang.
\newblock Gaittransformer: Multiple-temporal-scale transformer for cross-view gait recognition.
\newblock In {\em 2022 IEEE International Conference on Multimedia and Expo (ICME)}, pages 1--6, 2022.

\bibitem{dosovitskiy2020image}
Alexey Dosovitskiy, Lucas Beyer, Alexander Kolesnikov, Dirk Weissenborn, Xiaohua Zhai, Thomas Unterthiner, Mostafa Dehghani, Matthias Minderer, Georg Heigold, Sylvain Gelly, et~al.
\newblock An image is worth 16x16 words: Transformers for image recognition at scale.
\newblock {\em arXiv preprint arXiv:2010.11929}, 2020.

\bibitem{dou2022metagait}
Huanzhang Dou, Pengyi Zhang, Wei Su, Yunlong Yu, and Xi Li.
\newblock Metagait: Learning to learn an omni sample adaptive representation for gait recognition.
\newblock In {\em European Conference on Computer Vision}, pages 357--374. Springer, 2022.

\bibitem{Fan_2023_CVPR}
Chao Fan, Junhao Liang, Chuanfu Shen, Saihui Hou, Yongzhen Huang, and Shiqi Yu.
\newblock Opengait: Revisiting gait recognition towards better practicality.
\newblock In {\em Proceedings of the IEEE/CVF Conference on Computer Vision and Pattern Recognition (CVPR)}, pages 9707--9716, June 2023.

\bibitem{fan2020gaitpart}
Chao Fan, Yunjie Peng, Chunshui Cao, Xu Liu, Saihui Hou, Jiannan Chi, Yongzhen Huang, Qing Li, and Zhiqiang He.
\newblock Gaitpart: Temporal part-based model for gait recognition.
\newblock In {\em Proceedings of the IEEE/CVF conference on computer vision and pattern recognition}, pages 14225--14233, 2020.

\bibitem{Li_2023_CVPR}
Weijia Li, Saihui Hou, Chunjie Zhang, Chunshui Cao, Xu Liu, Yongzhen Huang, and Yao Zhao.
\newblock An in-depth exploration of person re-identification and gait recognition in cloth-changing conditions.
\newblock In {\em Proceedings of the IEEE/CVF Conference on Computer Vision and Pattern Recognition (CVPR)}, pages 13824--13833, June 2023.

\bibitem{li2020end}
Xiang Li, Yasushi Makihara, Chi Xu, Yasushi Yagi, Shiqi Yu, and Mingwu Ren.
\newblock End-to-end model-based gait recognition.
\newblock In {\em Proceedings of the Asian conference on computer vision}, 2020.

\bibitem{liao2020model}
Rijun Liao, Shiqi Yu, Weizhi An, and Yongzhen Huang.
\newblock A model-based gait recognition method with body pose and human prior knowledge.
\newblock {\em Pattern Recognition}, 98:107069, 2020.

\bibitem{lin2021gait}
Beibei Lin, Shunli Zhang, and Xin Yu.
\newblock Gait recognition via effective global-local feature representation and local temporal aggregation.
\newblock In {\em Proceedings of the IEEE/CVF International Conference on Computer Vision}, pages 14648--14656, 2021.

\bibitem{liu2021swin}
Ze Liu, Yutong Lin, Yue Cao, Han Hu, Yixuan Wei, Zheng Zhang, Stephen Lin, and Baining Guo.
\newblock Swin transformer: Hierarchical vision transformer using shifted windows.
\newblock In {\em Proceedings of the IEEE/CVF International Conference on Computer Vision}, pages 10012--10022, 2021.

\bibitem{liu2022video}
Ze Liu, Jia Ning, Yue Cao, Yixuan Wei, Zheng Zhang, Stephen Lin, and Han Hu.
\newblock Video swin transformer.
\newblock In {\em Proceedings of the IEEE/CVF Conference on Computer Vision and Pattern Recognition}, pages 3202--3211, 2022.

\bibitem{long2015fully}
Jonathan Long, Evan Shelhamer, and Trevor Darrell.
\newblock Fully convolutional networks for semantic segmentation.
\newblock In {\em Proceedings of the IEEE conference on computer vision and pattern recognition}, pages 3431--3440, 2015.

\bibitem{mogan2022gait}
Jashila~Nair Mogan, Chin~Poo Lee, Kian~Ming Lim, and Kalaiarasi~Sonai Muthu.
\newblock Gait-vit: Gait recognition with vision transformer.
\newblock {\em Sensors}, 22(19):7362, 2022.

\bibitem{nixon2006automatic}
Mark~S Nixon and John~N Carter.
\newblock Automatic recognition by gait.
\newblock {\em Proceedings of the IEEE}, 94(11):2013--2024, 2006.

\bibitem{pinvcic2022gait}
Domagoj Pin{\v{c}}i{\'c}, Diego Su{\v{s}}anj, and Kristijan Lenac.
\newblock Gait recognition with self-supervised learning of gait features based on vision transformers.
\newblock {\em Sensors}, 22(19):7140, 2022.

\bibitem{ren2015faster}
Shaoqing Ren, Kaiming He, Ross Girshick, and Jian Sun.
\newblock Faster r-cnn: Towards real-time object detection with region proposal networks.
\newblock {\em Advances in neural information processing systems}, 28, 2015.

\bibitem{Shen_2023_CVPR}
Chuanfu Shen, Chao Fan, Wei Wu, Rui Wang, George~Q. Huang, and Shiqi Yu.
\newblock Lidargait: Benchmarking 3d gait recognition with point clouds.
\newblock In {\em Proceedings of the IEEE/CVF Conference on Computer Vision and Pattern Recognition (CVPR)}, pages 1054--1063, June 2023.

\bibitem{szegedy2016rethinking}
Christian Szegedy, Vincent Vanhoucke, Sergey Ioffe, Jon Shlens, and Zbigniew Wojna.
\newblock Rethinking the inception architecture for computer vision.
\newblock In {\em Proceedings of the IEEE conference on computer vision and pattern recognition}, pages 2818--2826, 2016.

\bibitem{takemura2018multi}
Noriko Takemura, Yasushi Makihara, Daigo Muramatsu, Tomio Echigo, and Yasushi Yagi.
\newblock Multi-view large population gait dataset and its performance evaluation for cross-view gait recognition.
\newblock {\em IPSJ Transactions on Computer Vision and Applications}, 10(1):1--14, 2018.

\bibitem{vaswani2017attention}
Ashish Vaswani, Noam Shazeer, Niki Parmar, Jakob Uszkoreit, Llion Jones, Aidan~N Gomez, {\L}ukasz Kaiser, and Illia Polosukhin.
\newblock Attention is all you need.
\newblock {\em Advances in neural information processing systems}, 30, 2017.

\bibitem{wang2016temporal}
Limin Wang, Yuanjun Xiong, Zhe Wang, Yu Qiao, Dahua Lin, Xiaoou Tang, and Luc Van~Gool.
\newblock Temporal segment networks: Towards good practices for deep action recognition.
\newblock In {\em European conference on computer vision}, pages 20--36. Springer, 2016.

\bibitem{wu2016comprehensive}
Zifeng Wu, Yongzhen Huang, Liang Wang, Xiaogang Wang, and Tieniu Tan.
\newblock A comprehensive study on cross-view gait based human identification with deep cnns.
\newblock {\em IEEE transactions on pattern analysis and machine intelligence}, 39(2):209--226, 2016.

\bibitem{yu2006framework}
Shiqi Yu, Daoliang Tan, and Tieniu Tan.
\newblock A framework for evaluating the effect of view angle, clothing and carrying condition on gait recognition.
\newblock In {\em 18th International Conference on Pattern Recognition (ICPR'06)}, volume~4, pages 441--444. IEEE, 2006.

\bibitem{zheng2022gait}
Jinkai Zheng, Xinchen Liu, Wu Liu, Lingxiao He, Chenggang Yan, and Tao Mei.
\newblock Gait recognition in the wild with dense 3d representations and a benchmark.
\newblock In {\em Proceedings of the IEEE/CVF Conference on Computer Vision and Pattern Recognition}, pages 20228--20237, 2022.

\bibitem{zhu2021gait}
Zheng Zhu, Xianda Guo, Tian Yang, Junjie Huang, Jiankang Deng, Guan Huang, Dalong Du, Jiwen Lu, and Jie Zhou.
\newblock Gait recognition in the wild: A benchmark.
\newblock In {\em Proceedings of the IEEE/CVF international conference on computer vision}, pages 14789--14799, 2021.

\end{thebibliography}
}

\end{document}